\pdfoutput=1

\documentclass[11pt]{article}

\usepackage[]{ACL2023}

\usepackage{times}
\usepackage{latexsym}
\usepackage{multirow}
\usepackage{url}
\usepackage{latexsym}
\usepackage{balance}
\usepackage{bm}
\usepackage{amssymb}
\usepackage{amsmath}
\usepackage{url}
\usepackage{makecell}
\usepackage{multirow}
\usepackage{subfig}
\usepackage{graphicx} 
\usepackage{color}
\usepackage{colortbl}
\usepackage{textcomp}
\usepackage{CJK}
\usepackage{enumitem}
\usepackage{booktabs}
\usepackage{microtype}
\usepackage{arydshln}
\usepackage{amsfonts}
\usepackage{eufrak}
\usepackage[T1]{fontenc}

\usepackage[utf8]{inputenc}

\usepackage{microtype}
\usepackage{pifont}
\usepackage{inconsolata}

\title{Revisiting Token Dropping Strategy in Efficient BERT Pretraining}

\author{%
  Qihuang~Zhong$^{1}$,
  Liang~Ding$^{2}$,
  \textbf{Juhua~Liu}$^{3}$\thanks{~~Corresponding Authors: Juhua Liu (e-mail: liujuhua@whu.edu.cn), Bo Du (e-mail: dubo@whu.edu.cn)},
  Xuebo~Liu$^{4}$\\
  \textbf{Min~Zhang}$^{4}$, 
  \textbf{Bo~Du}$^{1*}$,
  \textbf{Dacheng~Tao}$^{5}$ \\
  \fontsize{9.0pt}{\baselineskip}\selectfont $^{1}$ National Engineering Research Center for Multimedia Software, Institute of Artificial Intelligence, School of Computer Science \\ 
  \fontsize{9.0pt}{\baselineskip}\selectfont  and Hubei Key Laboratory of Multimedia and Network Communication Engineering, Wuhan University, China \\
  \fontsize{9.0pt}{\baselineskip}\selectfont $^{2}$ JD~Explore~Academy, China \quad $^{3}$ Research Center for Graphic Communication, Printing and Packaging, \\
  \fontsize{9.0pt}{\baselineskip}\selectfont and Institute of Artificial Intelligence, Wuhan University, China   \\
  \fontsize{9.0pt}{\baselineskip}\selectfont $^{4}$ Institute of Computing and Intelligence, Harbin Institute of Technology, China  \quad $^{5}$ University of Sydney, Australia \\
   \fontsize{9.0pt}{\baselineskip}\selectfont \texttt{\{zhongqihuang, liujuhua, dubo\}@whu.edu.cn},
   \texttt{\{liangding.liam, dacheng.tao\}@gmail.com}
}

\begin{document}
\maketitle
\begin{abstract}
Token dropping is a recently-proposed strategy to speed up the pretraining of masked language models, such as BERT, by skipping the computation of a subset of the input tokens at several middle layers. It can effectively reduce the training time without degrading much performance on downstream tasks. However, we empirically find that token dropping is prone to a \textit{semantic loss} problem and falls short in handling semantic-intense tasks (\S\ref{sec:preliminaries}). Motivated by this, we propose a simple yet effective \textit{semantic-consistent} learning method (\textsc{ScTD}) to improve the token dropping. \textsc{ScTD} aims to encourage the model to learn how to preserve the semantic information in the representation space. Extensive experiments on 12 tasks show that, with the help of our \textsc{ScTD}, token dropping can achieve consistent and significant performance gains across all task types and model sizes. More encouragingly, \textsc{ScTD} saves up to 57\% of pretraining time and brings up to +1.56\% average improvement over the vanilla token dropping.
\end{abstract}
\section{Introduction}
\label{sec:intro}
Masked language models (MLMs), such as BERT~\cite{devlin2019bert} and its variants~\cite{liu2019roberta,he2020deberta,zhong2023se}\footnote{We refer to these models as BERT-style models.}, have achieved great success in a variety of natural language understanding (NLU) tasks. 
However, with the scaling of model size and corpus size, the pretraining of these BERT-style models becomes more computationally expensive and memory intensive~\cite{jiao2020tinybert,hou2022token}. Hence, it is crucial and green to speed up the training and reduce the computational overhead for BERT-style pretraining~\cite{zhang2020accelerating,schwartz2020green}. 

\begin{figure}[t]
    \includegraphics[width=0.48\textwidth]{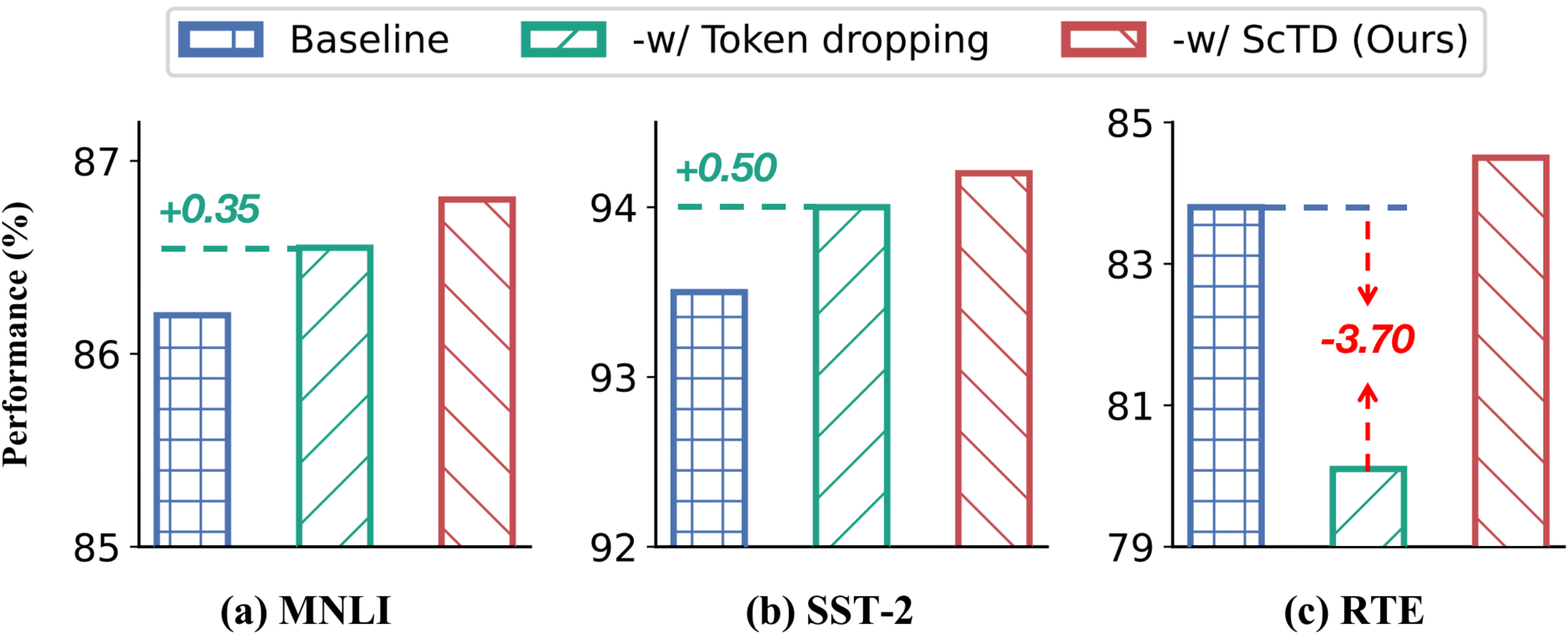} 
    \centering
    \caption{Performance of BERT$\rm_{\texttt{base}}$ on several downstream tasks. We see that: 1) Despite the remarkable performance on general tasks (\textit{i.e.}, MNLI and SST-2), token dropping leads to dramatically poor performance on the semantic-intense task (\textit{i.e.}, RTE). 2) Our \textsc{ScTD} achieves consistent performance gains among all tasks.}
    \label{fig:overall_result}
\end{figure}

To achieve this goal, various training-efficient approaches have been developed and summarized~\cite{shoeybi2019megatron,you2019large,zhang2020accelerating,shen2023efficient}. Among these efforts, a recently-proposed \textbf{token dropping}\footnote{We also refer to it as ``token drop'' in some cases.} strategy~\cite{hou2022token} has attracted increasing attention owing to its easy-to-implement algorithm and impressive efficiency (reducing the training cost by 25\% without much average performance dropping)~\cite{yao2022random,chiang2022recent}. Different from most previous works that focus on changing model architecture or optimization process, token dropping aims to improve training efficiency by dynamically skipping the compute of the redundant (unimportant) tokens that are less informative to the current training, at some middle layers of BERT during training. 
Although achieving a remarkable speedup, the performance improvement of token dropping is usually limited and unstable, compared to the baseline training scheme. More specifically, we empirically found that token dropping falls short in handling semantic-intense tasks, as shown in Figure~\ref{fig:overall_result}. This motivates us to explore and address the limitations of token dropping in this paper.

In light of the conventional wisdom that ``semantic information is mainly encoded in the BERT's intermediate and top layers''~\cite{jawahar2019does}, we suspected, apriori, that the corruption caused by the removal of unimportant tokens would break the sentence structure, and may easily lead to the semantic drift of sentence representations, as also observed in many similar scenarios~\cite{zhang2022instance,wang2021textflint}. To verify this conjecture, we conduct a series of preliminary analyses on a representative BERT model, and find that: 
\begin{itemize}
    \item[]\ding{182} The \textbf{\em training dynamics} of the token dropping show a significant semantic drift.
    \item[]\ding{183} The \textbf{\em representation} of a well-trained BERT with token dropping contains less semantics.
    \item[]\ding{184} The \textbf{\em downstream semantic-intense tasks} show a clear performance degradation.
\end{itemize}






Based on these observations, we can basically conclude that (one of) the limitation of token dropping is the \textit{semantic loss}\footnote{As we find that BERT models trained with token dropping are prone to losing some semantic-related polarities, e.g., less semantic knowledge in the dropped layers, we refer to this phenomenon as ``semantic loss" in the paper.} problem, which causes vulnerable and unstable training of BERT models.
To address this limitation, we propose a simple yet effective semantic-consistent learning method (referred to as \textbf{\textsc{ScTD}}) to improve token dropping. The principle of \textsc{ScTD} is to encourage the BERT to learn how to preserve the semantic information in the representation space. Specifically, \textsc{ScTD} first introduces two semantic constraints to align the semantic information of representations between baseline- and token dropping-based models, and then adopts a novel hybrid training approach to further improve the training efficiency.


We evaluate \textsc{ScTD} on a variety of benchmarks, including GLUE~\cite{wang2018glue}, SuperGLUE~\cite{wang2019superglue} and SQuAD v1/v2~\cite{rajpurkar2016squad,rajpurkar2018know}, upon two typical MLMs: BERT-\textsc{Base} and -\textsc{Large}. Results show that \textsc{ScTD} can not only bring consistent and significant improvements (up to +1.56\% average score among all tasks) into the token dropping strategy on both BERT models, but also alleviate the semantic loss problem. Moreover, compared to the standard BERT models, \textsc{ScTD} can also save up to 48\% of pretraining time while achieving comparable performance, and further achieve +1.42\% average gain for the same training iterations. 

To summarize, \textbf{our contributions} are as follows: 
\begin{itemize}
    \item Our study reveals the semantic loss problem in the token dropping strategy, which limits its performance on downstream tasks, especially on semantic-intense tasks.
    \item We propose a simple yet effective, plug-in-play approach (\textsc{ScTD}) to alleviate the semantic loss and further improve efficiency.
    \item Experiments show that \textsc{ScTD} outperforms the vanilla token dropping with up to +1.56\% average improvement and saves up to 57\% of pretraining time.
\end{itemize}

\section{Revisiting Token Dropping Strategy}
\label{sec:preliminaries}
In this section, we first review the background of token dropping strategy and then present the empirical analyses of this strategy in detail.

\begin{figure}[ht]
\centering
\includegraphics[width=0.47\textwidth]{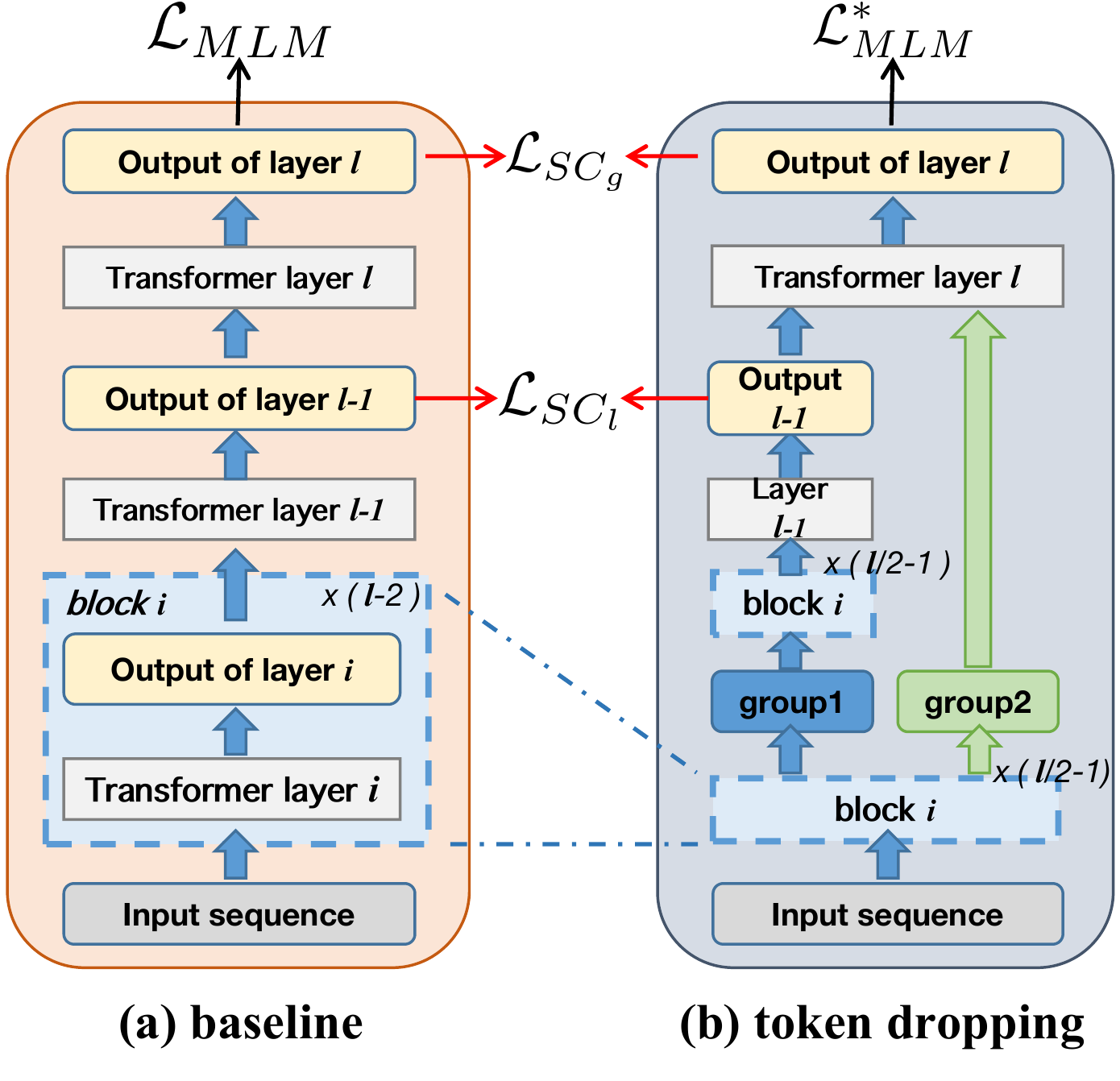}
\caption{Illustration of BERT-style models with (a) baseline training and (b) token dropping training. In (b), the ``\textcolor[rgb]{0.356,0.608,0.845}{\textbf{group1}}'' and ``\textcolor[rgb]{0.651,0.804,0.553}{\textbf{group2}}'' denote the important and unimportant (skipped) tokens, respectively. The $\mathcal{L}_{SC_l}$ and $\mathcal{L}_{SC_g}$ (in \textcolor{red}{red arrows}) refer to the semantic-align objectives used in our \textsc{ScTD}.}
\label{fig:method1}
\end{figure}

\subsection{Preliminaries}
Suppose that we focus on pretraining the BERT with $l$ transformer layers. Let $L_i$ denote the $i$-th ($i \in \{1,...,l\}$) layer, $X_i \in \mathbb{R}^{s_i \times d}$ be the output tensors of $i$-th layer, where $s_i$ is the sequence length of $i$-th layer and $d$ is the hidden size. Notably, $X_0$ denotes the input (after embedding) of the model. 
For the baseline training process (as illustrated in Figure~\ref{fig:method1} \textbf{(a)}), full-sequence tokens will be sequentially fed into all layers, \textit{i.e.}, $s_0=s_1=...=s_l$. In this way, we can obtain the final output tensors $X_l$ of $l$-th layer, and then use a cross-entropy loss to optimize the training process as follow:
\begin{align}
    \mathcal{L}_{MLM}= \mathbb{E} \left( - \sum \log P(Y|X_l) \right),
    \label{mlm_loss}
\end{align}
where $Y$ denotes the ground-truths.

For token dropping (as illustrated in Figure~\ref{fig:method1} \textbf{(b)}), different from the full-sequence training, the training of a subset of unimportant tokens in middle layers will be skipped\footnote{\citet{hou2022token} state that such a process would not only hardly damage the effect of pretraining, but also reduce the computation costs.}. In practice, for stable training, token dropping follows the full-sequence training at several first layers (\textit{i.e.}, from 1-th layer to $(l/2-1)$-th layer). Then, it uses several importance scores/metrics to determine the dropped tokens and divides tokens into two groups, where we denote the ``group1'' as important tokens and ``group2'' as unimportant (dropped) tokens. The group1 tokens will be fed into later layers (\textit{i.e.}, from $(l/2-1)$-th layer to $(l-1)$-th layer), while the computations of the group2 tokens are skipped. Lastly, all tokens are merged before the last layer and then are used to obtain the final outputs\footnote{To distinguish from final outputs $X_l$ of baseline training, we denote it as $\Tilde{X}_l$.} $\Tilde{X}_l$. The loss function of token dropping is similar to Eq.~\ref{mlm_loss}, and we refer to it as $\mathcal{L}^*_{MLM}$.

\subsection{Empirical Analyses}
In this part, to verify whether removing the unimportant tokens will cause the loss of semantic information and thus hinder the performance of token dropping, we conduct systematic analyses from three aspects: 1) \textit{revealing the semantic drift problem during \textbf{training dynamics}}; 2) \textit{probing the \textbf{representation} of a well-trained model with token dropping}; 3) \textit{evaluating the \textbf{downstream performance} on semantic-intense tasks}. In practice, for comparison, we pre-train the representative BERT$\rm_{\texttt{base}}$ models with baseline training scheme and token dropping, respectively. Through the above analyses, we empirically observe that:

\paragraph{\ding{182} The training dynamics of the token dropping show a significant semantic drift.}
As suspected in \S\ref{sec:intro}, the corruption caused by the removal of several tokens would break the sentence structure, thus leading to semantic drift. Here, we verify this conjecture by quantitatively estimating the loss of semantic information contained in the corrupted sentence. For measuring the semantic information, we first adopt the off-the-shelf Sentence-BERT~\cite{reimers-2019-sentence-bert} to capture the semantic representations. Then, suppose that the original sentence (without any corruption, such as masking or token dropping) contains full semantic information, we refer to the cosine similarity between semantic representations of the corrupted and original sentences as a metric to measure the semantic drift in the corrupted sentence.

\begin{figure}[ht]
\centering
\includegraphics[width=0.47\textwidth]{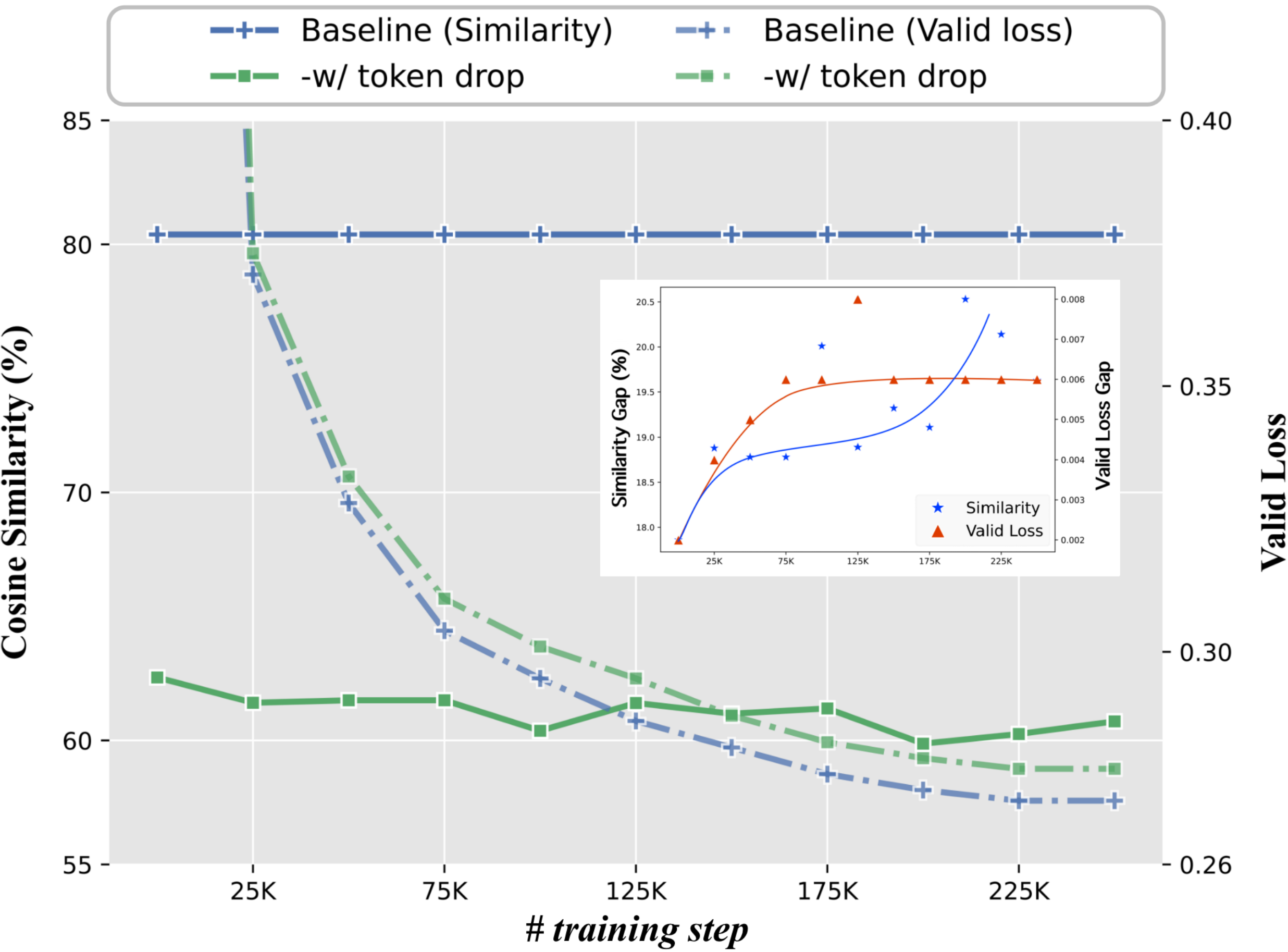}
\caption{The comparison of similarity and validation curves between baseline and token dropping on BERT$\rm_{\texttt{base}}$ pretraining. The left y-axis is the cosine similarity between corrupted- (in baseline and token dropping settings, respectively) and original sentences, while the right y-axis is the validation results. The similarity and validation gaps are illustrated in the inserted figure.}
\label{fig:pre_analysis1}
\end{figure}

\noindent In practice, given some sentences randomly sampled from training data, we follow the above process and measure the (average) semantic drift during the baseline/token dropping training dynamics, respectively. For reference, we also report the validation results and illustrate all results in Figure~\ref{fig:pre_analysis1}. It can be found that: compared to baseline training, \textit{i)} sentence semantics in token dropping drifts more from the original semantics; \textit{ii)} token dropping hinders the full learning of BERT, especially in the middle and later training stages (after 75K steps). To have a closer look, we show the similarity and validation gaps between both settings in the inserted figure of Figure~\ref{fig:pre_analysis1}. As seen, with the training going on, both gaps have a similar tendency to increase\footnote{The curve of validation gap tends to flatten in the later training stage, as both models are going to converge.}, especially at the beginning of training. In general, these analyses indicate that \textit{there is a significant semantic drift during training dynamics of token dropping, which shows a correlation with the performance drop of token dropping.}

\begin{figure}[h]
\centering
\includegraphics[width=0.47\textwidth]{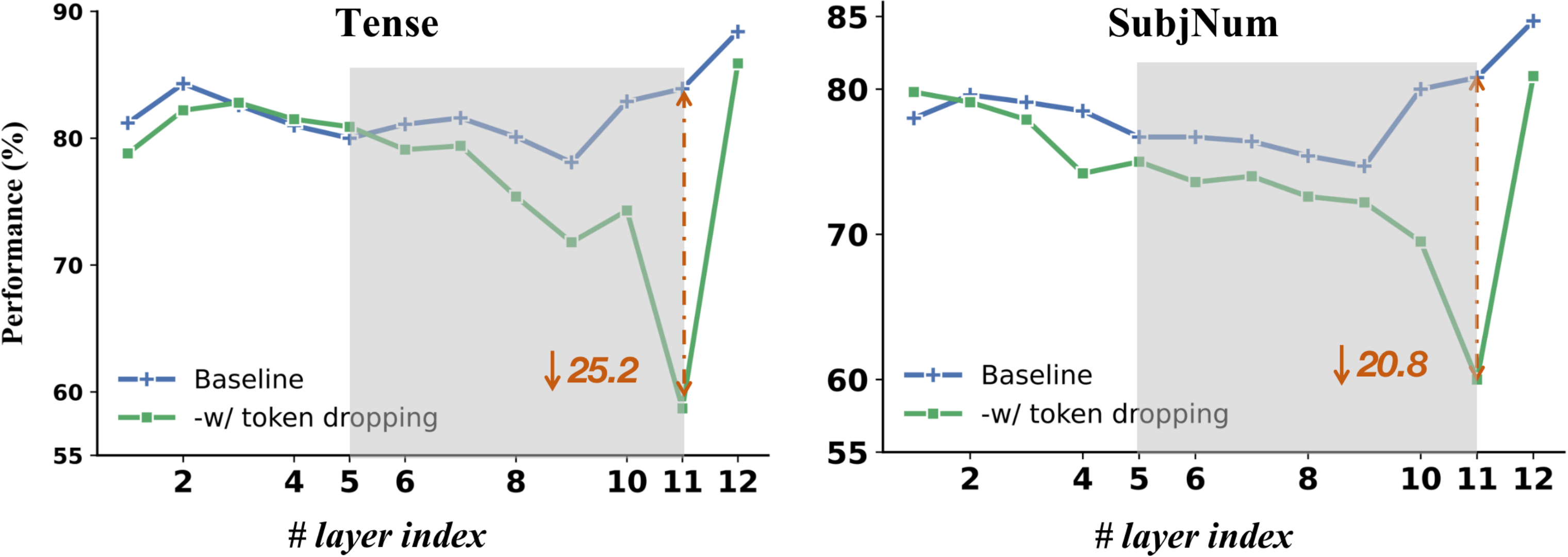}
\caption{The comparison of semantic information between baseline and token dropping on different BERT$\rm_{\texttt{base}}$ layers. We see that, for token dropping, as the number of dropped layers (from layer 5 to layer 11, illustrated in shadow areas) increases, the semantic information saved by the model is significantly reduced.}
\label{fig:pre_analysis2}
\end{figure}

\paragraph{\ding{183} The representation of a well-trained BERT with token dropping contains less semantics.}
In addition to the analysis during training dynamics, we then investigate the semantic properties of well-trained models. Specifically, following many prior works~\cite{conneau2018you,jawahar2019does,ding2020context,zhong2022e2s2}, we perform several semantic-aware probing tasks on the sentence representations at different layers. Taking the \textbf{Tense} and subject number (\textbf{SubjNum}) tasks as examples, we provide the comparison of semantic information between baseline and token dropping at different layers in Figure~\ref{fig:pre_analysis2}.

\noindent We observe that there is more semantic information in the top layers (from layer 9 to layer 12) of BERT trained with the baseline scheme, which is similar to the finding of~\citet{jawahar2019does}. However, when using the token dropping, the semantic information contained in BERT tends to decrease in the dropped layers (from layer 5 to layer 11). The semantic information of token dropping at 11-th layer drops dramatically, which is much lower (up to 25.2 points) than that of baseline. Moreover, due to the vulnerable and unstable training, the final representation in token dropping at the last layer is also sub-optimal. These results basically prove that \textit{the semantic drift of token dropping damages the semantic learning ability of BERT}.


\paragraph{\ding{184} The downstream semantic-intense tasks show a clear performance degradation.}
The aforementioned analyses mainly focus on interpreting the semantic properties of models. Here, we further evaluate the downstream performance of token dropping. Specifically, several representative semantic-intense\footnote{We chose tasks based on whether they require semantic-related information to solve. For instance, we included MRPC~\cite{dolan2005automatically}, a task that predicts if two sentences are semantically equivalent.} tasks are used, including OntoNotes 5.0~\cite{weischedel2013ontonotes} (Onto. for short), CoNLL03~\cite{sang2003introduction}, MRPC~\cite{dolan2005automatically} and SICK-Relatedness~\cite{marelli2014sick} (SICK-R for short). Notably, for Onto. and CoNLL03, we report the few-shot (32-shot) performance to enlarge the performance difference between different models. We measure the development performance of each task using its corresponding evaluation metrics, and report the contrastive results in Table~\ref{tab:pre_analysis3}.

\begin{table}[ht]
\centering
\scalebox{0.78}{
\begin{tabular}{lccccc}
\toprule
\multicolumn{1}{c}{\multirow{2}{*}{Method}} & Onto. & CoNLL03 & MRPC & SICK-R &\multicolumn{1}{c}{\multirow{2}{*}{\underline{Avg.}}} \\ \cmidrule(lr){2-5}
\multicolumn{1}{c}{} & \textit{F1} & \textit{F1} & \textit{Acc.} & \textit{Spear.}  \\ \midrule
Baseline & 30.16 & 54.48 & 86.80 &69.08 &\underline{60.13} \\
token drop & 27.49 & 53.73& 85.50 & 66.16 &\underline{58.22}  \\ \hdashline
\multicolumn{1}{c}{$\Delta$ ($\downarrow$)} & \textcolor{red}{\textbf{-2.67}} & \textcolor{red}{\textbf{-0.75}} & \textcolor{red}{\textbf{-1.30}} & \textcolor{red}{\textbf{-2.92}} &\underline{\textcolor{red}{\textbf{-1.91}}} \\
\bottomrule
\end{tabular}
}
\caption{Experimental results of BERT$\rm_{\texttt{base}}$ trained with different methods on several semantic-intense tasks. We observe that token dropping strategy leads to poor performance among all these tasks.}
\label{tab:pre_analysis3}
\end{table}

\noindent As seen, there is a great discrepancy between the downstream performance of baseline and token dropping. Overall, token dropping consistently under-performs the baseline with an average 1.91\% performance drop, among all these semantic-intense tasks. Specifically, as for SICK-R (usually used to measure the semantic textual similarity), token dropping performs much worse (up to $\downarrow$2.92) than the baseline. These results indicate that, \textit{due to the semantic drift, BERT with token dropping falls short in handling the semantic-intense tasks}.
\section{Improving Token Dropping with Semantic-Consistent Learning}
\label{sec:method}
Based on the observations in \S\ref{sec:preliminaries}, we recognize that it is essential to alleviate the side effect (i.e., \textit{semantic loss} problem) of token dropping. To achieve this goal, we propose a simple yet effective semantic-consistent learning (\textsc{ScTD}) framework Specifically, our \textsc{ScTD} adopts two key techniques as follows:

\begin{figure}[ht]
\centering
\includegraphics[width=0.47\textwidth]{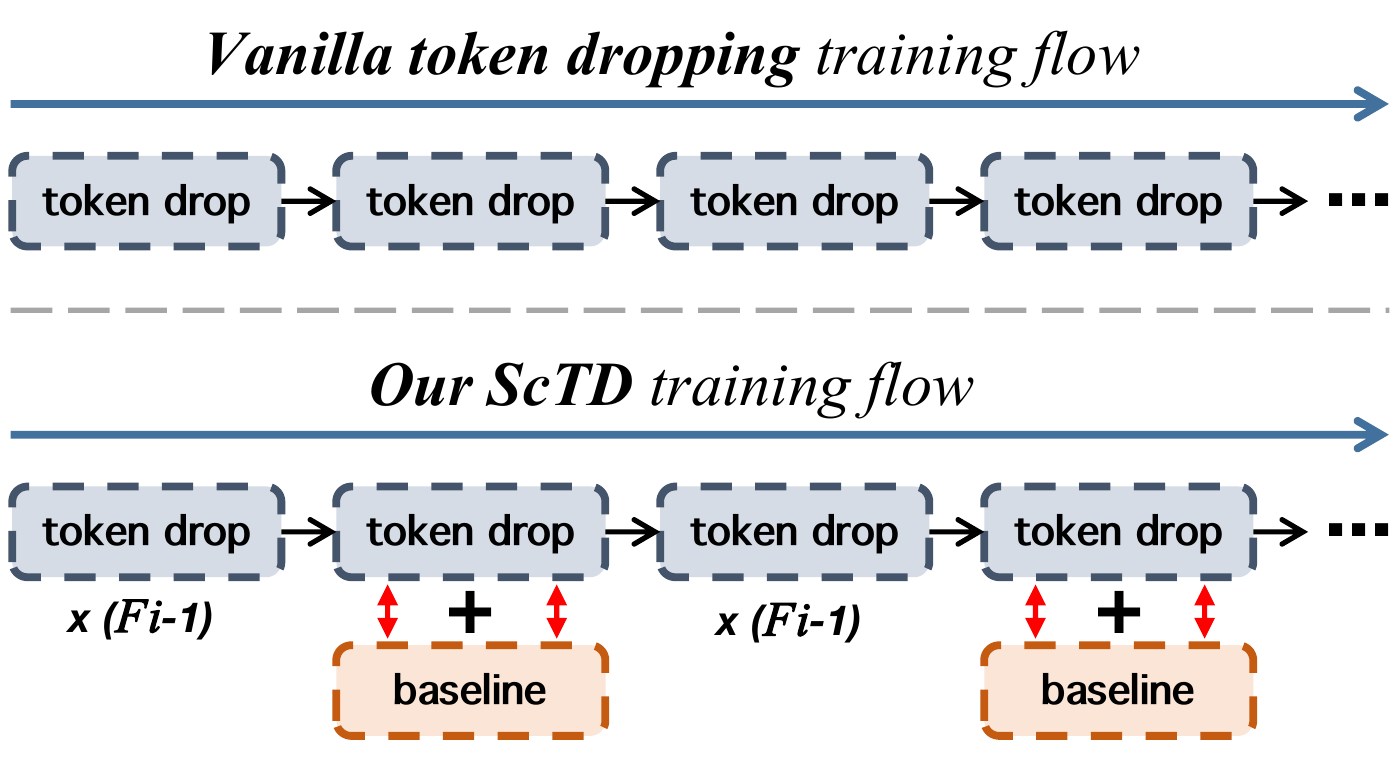}
\caption{The comparison of training flow between the vanilla token dropping and our \textsc{ScTD}. The ``token drop'' and ``baseline'' modules refer to the corresponding training processes in Figure~\ref{fig:method1}. For \textsc{ScTD}, ``$\times (F_i-1)$'' means repeating the token dropping process multiple times, where $F_i$ is a fixed interval.}
\label{fig:method2}
\end{figure}

\paragraph{Semantic-Consistent Learning.} The principle of our \textsc{ScTD} is to encourage the model to preserve the semantic information in the representation space. 
Inspired by the success of knowledge distillation~\cite{hinton2015distilling,xu2020knowledge}, \textsc{ScTD} refers to the model with baseline training (containing more semantic information) as the teacher to guide the training of students (i.e., model trained with token dropping). Considering that it is usually unavailable to obtain the pre-trained teacher model, we hereby recast it as a self-distillation process~\cite{zhang2020self,ding2021understanding}. Given the same input $X_0$, we input $X_0$ into the model to perform twice forward-propagation processes, where one is for token dropping and the other is for baseline training. The outputs of baseline training ($X_l$) are used as the teacher distributions to teach the student (outputs $\Tilde{X}_l$ of token dropping). As such, the student can learn how to align the semantic information with the teacher. More specifically, \textsc{ScTD} introduces two semantic constraints in a local-to-global manner (as illustrated in Figure~\ref{fig:method1}). For the global one, we use the KL divergence to constrain the \textit{global-level} semantic distributions of baseline- and token-dropping-based models at the last $l$-th layer, as follows:
\begin{align}
    \mathcal{L}_{SC_g}= {\bf KL} \left (p(X_l)||p(\Tilde{X}_l) \right ),
    \label{scg_loss}
\end{align}
where $p(X_l)$ and $p(\Tilde{X}_l)$ denote the corresponding distributions respectively. On the other hand, in slight of the finding that semantic loss is the most significant in the penultimate layer ($l-1$) in token dropping setting (Figure~\ref{fig:pre_analysis2}), we further construct a \textit{local-level} semantic constraint at the $(l-1)$-th layer, which is similar to Eq.~\ref{scg_loss}:
\begin{align}
    \mathcal{L}_{SC_l}= {\bf KL}\left (p(X_{l-1})||p(\Tilde{X}_{l-1}) \right ).
    \label{scl_loss}
\end{align}

\paragraph{Hybrid Training.} 
Since the semantic-consistent learning process requires twice forward/back-propagation, \textsc{ScTD} would introduce much computational overhead, leading to inefficiency. To overcome this issue, \textsc{ScTD} adopts a novel hybrid training strategy, as illustrated in Figure~\ref{fig:method2}. Specifically, instead of using the semantic-consistent learning method throughout the training, \textsc{ScTD} basically follows the vanilla token dropping and adopts the semantic-consistent training intermittently. As such, \textsc{ScTD} can combine the advantages of semantic-consistent learning (\textit{effectiveness}) and token dropping (\textit{efficiency}). Let $Fi$ be a fixed interval, \textsc{ScTD} first performs the vanilla token dropping training $(Fi-1)$ times and then performs once the semantic-consistent training. The overall training objective of \textsc{ScTD} can be formulated as:

\begin{equation}
    \mathcal{L}_{all}=
    \begin{cases}
    \begin{aligned}
    &\frac{1}{2}\mathcal{L}^*_{MLM}+\frac{1}{2}\mathcal{L}_{MLM} \\
    &+\lambda*(\mathcal{L}_{SC_g}+\mathcal{L}_{SC_l})
    \end{aligned}, &t\bmod Fi = 0 \vspace{2ex} \\  
    \mathcal{L}^*_{MLM}, &t\bmod Fi \neq 0
    \label{sc_loss}
\end{cases}
\end{equation}
where $t$ denotes the index of training iterations and $\lambda$ is a weight factor to balance the different objectives, which is empirically\footnote{The detailed analysis can be found in \S\ref{sec:ablation}.} set as 0.05.
\begin{table*}[]
\centering
\scalebox{0.88}{
\begin{tabular}{lccccccccccc}
\toprule
\multicolumn{1}{c}{\multirow{2}{*}{Method}} & \multicolumn{1}{c}{Budget} & CoLA & \multicolumn{2}{c}{MRPC} & \multicolumn{2}{c}{STS-B} & RTE & \multicolumn{2}{c}{MNLI} & SST-2 & GLUE \\ \cmidrule(lr){3-12}
\multicolumn{1}{c}{} & \multicolumn{1}{c}{\textit{hours}} & \textit{Mcc.} & \textit{Acc.} & \textit{F1} & \textit{Pear.} & \textit{Spea.} & \textit{Acc.} & \textit{m.} & \textit{mm.} & \textit{Acc.} & \textit{\underline{Avg.}} \\ \midrule
\multicolumn{12}{c}{BERT$\rm_{\texttt{large}}$} \\ \midrule
Baseline (\textit{250K}) & 34.35 & 61.3 & 90.0 & 92.7 & \textbf{90.2} & \textbf{89.9} & 83.8 & 86.3 & 86.1 &93.5 & \underline{84.37} \\
token drop (\textit{250K}) & 27.33 (-20\%) & 64.3 & 88.0 & 91.4 & 89.7 & 89.5 & 80.1 & 86.8 & 86.3 &94.0 & \underline{84.04} \\ \hdashline
-w/ \textsc{ScTD} (\textit{100K}) & 11.83 (-66\%) & 62.3 & 89.2 & 92.2 & 89.9 & 89.7 & 80.9 & 85.1 & 84.8 &93.0 & \underline{83.61} \\
-w/ \textsc{ScTD} (\textit{160K}) & 17.75 (-48\%) & \textbf{65.8} & 88.7 & 91.8 & 89.9 & 89.7 & 81.2 & 86.4 & 86.1 &94.0 & \underline{84.55} \\
-w/ \textsc{ScTD} (\textit{250K}) & 29.54 (-14\%) & 65.6 & \textbf{91.4} & \textbf{93.8} & \textbf{90.2} & \textbf{89.9} & \textbf{84.5} & \textbf{87.1} & \textbf{86.5} &\textbf{94.2} & \underline{\textbf{85.63}} \\  \midrule
\multicolumn{12}{c}{BERT$\rm_{\texttt{base}}$} \\ \midrule
Baseline (\textit{250K}) &15.17  & 56.0 & 86.8 & 90.1 & \textbf{89.0} & \textbf{88.8} & 77.6 & 83.3 & 83.5 &\textbf{92.3} & \underline{81.11} \\
token drop (\textit{250K}) &12.92 (-15\%)  & 54.1 & 85.5 & 89.6 & 87.8 & 87.8 & 77.6 & 83.4 & 83.3 &91.7 & \underline{80.35} \\
\hdashline
-w/ \textsc{ScTD} (\textit{100K}) &5.51 (-64\%)  & 55.4 & \textbf{87.3} & \textbf{91.1} & 88.4 & 88.3 & 76.9 & 82.2 & 82.4 & 91.4 & \underline{80.59} \\
-w/ \textsc{ScTD} (\textit{160K}) &8.79 (-42\%)  & 58.1 & 87.0 & 90.7 & 88.1 & 88.0 & 78.7 & 83.4 & 83.3 & 90.6 & \underline{81.28} \\
-w/ \textsc{ScTD} (\textit{250K}) &13.78 (-9.2\%)  & \textbf{58.8} & 86.8 & 90.5 & 88.2 & 88.1 & \textbf{79.4} & \textbf{83.8} & \textbf{83.6} & 91.6 & \underline{\textbf{81.72}} \\
\bottomrule
\end{tabular}
}
\caption{Experimental results (dev scores) of BERT$\rm_{\texttt{large}}$ and BERT$\rm_{\texttt{base}}$ trained with different methods on the GLUE benchmark. Average scores on all tasks are \underline{underlined}. The best results are in \textbf{bold}. We see that our \textsc{ScTD} improves the performance and training efficiency of token drop strategy across all task types and model sizes.}
\label{tab_main}
\end{table*}

\section{Evaluation}
\label{sec:experiments}
\subsection{Setup}
\paragraph{Downstream Tasks}
To investigate the effectiveness and universality of \textsc{ScTD}, we follow many previous studies~\cite{zhong2022panda,zhong2022toward} and conduct extensive experiments on various NLU tasks, covering a diversity of tasks from GLUE~\cite{wang2018glue}, SuperGLUE~\cite{wang2019superglue} and SQuAD benchmarks. Specifically, three semantic-intense tasks (MRPC~\cite{dolan2005automatically}, STS-B~\cite{cer2017semeval} and RTE~\cite{giampiccolo2007third}), five question answering tasks (BoolQ~\cite{clark2019boolq}, COPA~\cite{roemmele2011choice}, MultiRC~\cite{khashabi2018looking}, SQuAD-v1~\cite{rajpurkar2016squad} and -v2~\cite{rajpurkar2018know}), two natural language inference tasks (MNLI~\cite{williams2018broad} and CB~\cite{de2019commitmentbank}), and two others (CoLA~\cite{warstadt2019neural} and SST-2~\cite{socher2013recursive}) are used. For evaluation, we report the performance with Accuracy (``\textit{Acc.}'') metric for most tasks, except the Pearson and Spearman correlation (``\textit{Pear./Spea.}'') for STS-B, the Matthew correlation (``\textit{Mcc.}'') for CoLA, the F1 score for MultiRC, and the Exact Match (``\textit{EM}'') scores for SQuAD v1/v2.
We report the averaged results over 5 random seeds to avoid stochasticity.
The details of all tasks and datasets are shown in Appendix~\ref{appendix_data}.

\paragraph{Hyper-parameters}
For pretraining, we train the BRET-\textsc{Base} and -\textsc{Large} models with different methods\footnote{Following~\citet{hou2022token}, we implement the token dropping and our approach under the same settings, \textit{e.g.}, dropping 50\% of the tokens.} from scratch. We basically follow the original paper~\cite{devlin2019bert} (\textit{e.g.}, the same pretraining corpus), except that we do not use the next sentence prediction (NSP) objective, as suggested in~\cite{liu2019roberta}. In practice, we train each model for 250K steps, with a batch size of 1024 and a peak learning rate of 2e-4. For fine-tuning, the learning rate is selected in \{1e-5, 2e-5, 3e-5, 5e-5\}, while the batch size is in \{12, 16, 32\} depending on tasks. The maximum length of input sentence is 384 for SQuAD v1/v2 and 256/512 for other tasks. The detailed hyper-parameters for fine-tuning are provided in Appendix~\ref{appendix_parameters}. 
We use AdamW~\cite{loshchilov2018decoupled} as the optimizer for both pretraining and fine-tuning processes. All experiments are conducted on NVIDIA A100 (40GB) GPUs. 

\begin{table}[]
\centering
\scalebox{0.72}{
\begin{tabular}{lcccccc}
\toprule
\multicolumn{1}{c}{\multirow{2}{*}{Method}} & Boolq & CB & MultiRC & COPA & \multicolumn{1}{c}{SQ-v1} & \multicolumn{1}{c}{SQ-v2} \\ \cmidrule(lr){2-5} \cmidrule(lr){6-7}
\multicolumn{1}{c}{} & \textit{Acc.} & \textit{Acc.} & \textit{F1} & \textit{Acc.} & \textit{EM} & \textit{EM} \\ \midrule
\multicolumn{7}{c}{BERT$\rm_{\texttt{large}}$} \\ \midrule
Baseline & 78.1 & 91.1 & 70.3 & \textbf{72.0} &85.53 &79.16 \\
token drop & 79.9 & 91.1 & \textbf{72.8} & 68.0 &86.35 &81.50  \\
-w/ \textsc{\textsc{ScTD}} & \textbf{79.7} & \textbf{92.9} & \textbf{72.8} & \textbf{72.0} &\textbf{86.54} & \textbf{81.67} \\ \midrule
\multicolumn{7}{c}{BERT$\rm_{\texttt{base}}$} \\ \midrule
Baseline &  \textbf{74.4} & 83.9 & 68.1 & 63.0 &81.97 &72.18  \\
token drop &  73.0 & 83.9 & 67.7 & 64.0 &81.67 &72.68  \\
-w/ \textsc{\textsc{ScTD}} &  73.8 & \textbf{87.5} & \textbf{68.9} & \textbf{68.0} &\textbf{82.47} &\textbf{72.79}  \\
\bottomrule
\end{tabular}
}
\caption{Experimental results of BERT$\rm_{\texttt{large}}$ and BERT$\rm_{\texttt{base}}$ trained with different methods on the SuperGLUE~\cite{wang2019superglue}  benchmark and SQuAD~\cite{rajpurkar2016squad} (SQ for short) tasks. We see that our \textsc{\textsc{ScTD}} achieves consistent and significant improvements on SuperGLUE and SQuAD tasks as well.}
\label{tab_main2}
\end{table}

\begin{figure}[t]
    \centering
    \includegraphics[width=0.45\textwidth]{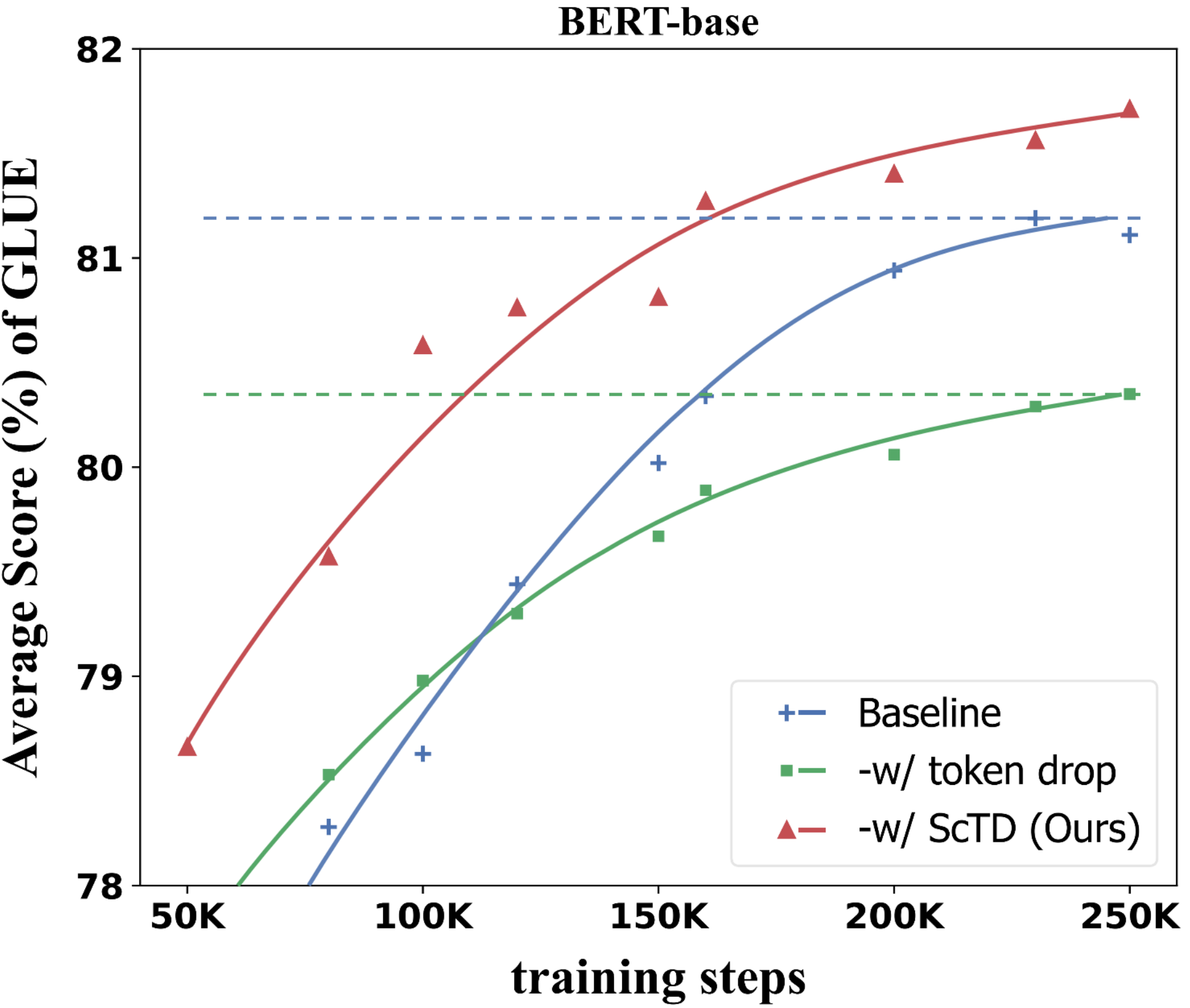}
    \caption{Average scores (\%) on GLUE benchmark of BERT$\rm_{\texttt{base}}$ models trained with different methods for the full pretraining process. Our method achieves comparable performance with baseline at 150K training steps. }
    \label{fig_training}
\end{figure}

\subsection{Compared Results}
Results of GLUE are shown in Table~\ref{tab_main}, while those of SuperGLUE and SQuAD are in Table~\ref{tab_main2}. Based on these results, we can find that:

\paragraph{\textsc{ScTD} consistently improves performance on all types of tasks.} First, results on the semantic-intense tasks (MRPC, STS-B and RTE) show that \textsc{ScTD} effectively alleviates the semantic loss problem of token dropping. Specifically, for the RTE task, \textsc{ScTD} brings significant improvement (up to +3.4\%) against the vanilla token dropping, and even outperforms the full-sequence training baseline. On the other hand, we observe that \textsc{ScTD} is also beneficial to the other general tasks (\textit{e.g.}, question answering). With the help of \textsc{ScTD}, token dropping strategy achieves up to +1.56\% average gains among all types of tasks, proving the effectiveness and universality of \textsc{ScTD}.

\paragraph{\textsc{ScTD} improves performance on both model sizes.} Extensive results show that \textsc{ScTD} works well on both Large and Base BERT models. Specifically, compared to the vanilla token dropping, \textsc{ScTD} brings +1.59\% and +1.37\% average gains on GLUE tasks, respectively. Results on the other tasks also show a similar phenomenon. Thus, we could recommend our \textsc{ScTD} to speed up the training of all discriminative MLMs regardless of the regime in model capacity.

\paragraph{\textsc{ScTD} effectively improves the training efficiency.} Results in Table~\ref{tab_main} show that, with our \textsc{ScTD}, BERT models can achieve comparable or even better performance with much fewer training steps, \textit{i.e.}, improving the training efficiency\footnote{While the semantic-consistent learning process in \textsc{ScTD} will introduce extra computation overhead, \textsc{ScTD} performs much better in terms of training efficiency. That is, the relatively little computation overhead is acceptable.}. Specifically, compared to the full training (250K steps) BERT models, \textsc{ScTD} can save up to 48\% pretraining time while achieving comparable performance. We attribute it to the higher data efficiency, since \textsc{ScTD} not only takes full advantage of the token dropping's ability to learn important words but also alleviates the semantic loss problem in the token dropping. This can be further proved by the illustration of Figure~\ref{fig_training}, as \textsc{ScTD} always shows better performance against the other counterparts during the training dynamics. Furthermore, when training with the same iterations, our \textsc{ScTD} can even outperform the standard BERT by a clear margin. We attribute this to the regularization effect of token dropping\footnote{BERT-style PLMs are often over-parameterized and prone to overfitting. Using regularization methods like token dropping and LayerDrop~\cite{fanreducing} during training can improve model generalization and even boost performance.}.

\begin{table}[]
\centering
\scalebox{0.8}{
\begin{tabular}{cccccc}
\toprule
 \multirow{2}{*}{$\mathcal{L}_{MLM}$} & \multirow{2}{*}{$\mathcal{L}_{SC_l}$} & \multirow{2}{*}{$\mathcal{L}_{SC_g}$} & GLUE &SGLUE &SQuAD \\ \cmidrule{4-6}
&  &  & \textit{Avg.} & \textit{Avg.} & \textit{Avg.} \\ \midrule
\multicolumn{3}{l}{Baseline} & 77.73 &69.11 &74.15 \\
\multicolumn{3}{l}{token drop} & 76.58 &68.01 &72.28 \\ \midrule
\multicolumn{6}{l}{-w/ \textsc{ScTD} (Ours)} \\ 
\checkmark &  & & 78.30 &68.73  &75.56  \\
 & \checkmark &  & 78.06 &69.49  &75.66  \\
 &  & \checkmark & 79.27 &68.64  &75.80  \\
\checkmark & \checkmark &  & 78.51 &69.64  &75.51  \\
\checkmark &  & \checkmark & 79.26 &69.32  &75.59  \\
 & \checkmark & \checkmark & 79.36 &69.89  &75.91  \\
\checkmark & \checkmark & \checkmark & \textbf{79.58} &\textbf{70.29}  &\textbf{76.01}  \\
\bottomrule
\end{tabular}
}
\caption{Ablation study on different training objectives ($\{\mathcal{L}_{MLM},\mathcal{L}_{SC_l},\mathcal{L}_{SC_g}\}$) introduced in our \textsc{ScTD}.}
\label{tab:ablation_loss}
\end{table}

\subsection{Ablation Study}
\label{sec:ablation}
We evaluate the impact of each component of our \textsc{ScTD}, including \textit{i}) semantic-consistent learning objectives, \textit{ii}) coefficient $\lambda$ and \textit{iii}) fixed interval $Fi$ in the hybrid training process. Notably, due to the limited computational budget, we conduct experiments on the BERT$\rm_{\texttt{large}}$ models trained with different methods for 5 epochs (35K steps).

\paragraph{Impact of different training objectives.}
As shown in \S\ref{sec:method}, in addition to the original MLM objective $\mathcal{L}^*_{MLM}$ of token dropping, we introduce several extra training objectives ($\mathcal{L}_{MLM},\mathcal{L}_{SC_l},\mathcal{L}_{SC_g}\}$) to align the semantic information. Here, we conduct experiments to analyze the impact of different objectives and show the results in Table~\ref{tab:ablation_loss}. It can be seen that all objectives are beneficial to our \textsc{ScTD}, where the $\mathcal{L}_{SC_g}$ is the most helpful. This indicates the semantic alignment in the global-level representation space is more critical. Also, we can observe that the combination of all objectives performs best, thus leaving as the default setting.

\begin{figure}[t]
    \centering
    \includegraphics[width=0.42\textwidth]{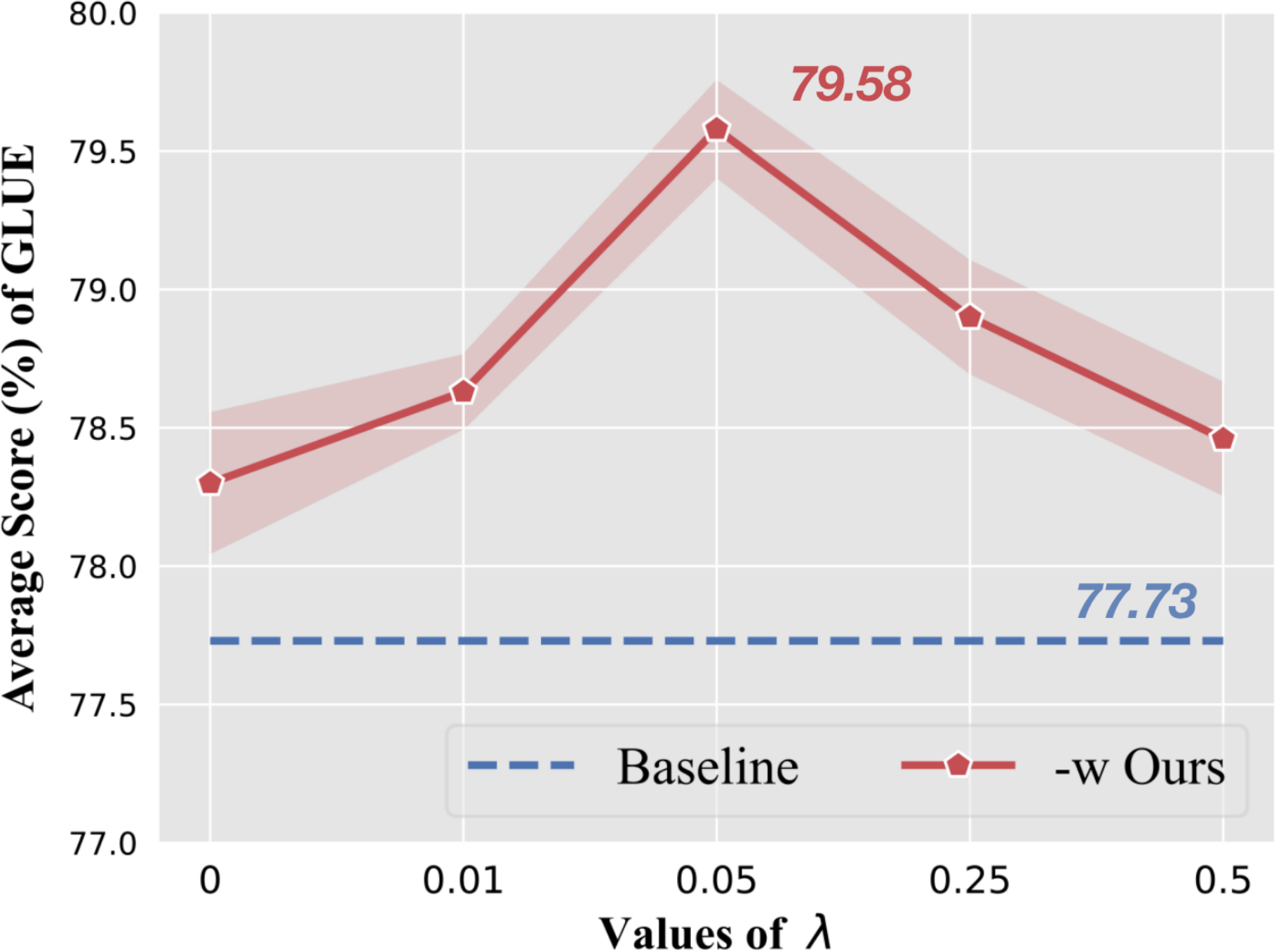}
    \caption{Parameter analysis of $\lambda$ on BERT$\rm_{\texttt{large}}$. }
    \label{fig:ablation2}
\end{figure}
\paragraph{Impact of Coefficient $\lambda$.}
The factor $\lambda$ in Eq.~\ref{sc_loss}, which is used to balance different objectives, is an important hyper-parameters. In this study, we analyze its influence by evaluating the performance with different $\lambda$ spanning \{0, 0.01, 0.05, 0.25, 0.5\} on several GLUE tasks. Figure~\ref{fig:ablation2} illustrates the average results. Compared with the baseline, our \textsc{ScTD} consistently brings improvements across all ratios of $\lambda$, basically indicating that the performance of \textsc{ScTD} is not sensitive to $\lambda$. More specifically, the case of $\lambda = 0.05$ performs best, and we thereby use this setting in our experiments.

\begin{table}[]
\centering
\scalebox{0.85}{
\begin{tabular}{lccc}
\toprule
\multicolumn{1}{c}{\multirow{2}{*}{Method}} & Budget & GLUE & SQuAD \\ \cmidrule{2-4}
\multicolumn{1}{c}{} & \textit{training time (hours)} & \textit{Avg.} & \textit{Avg.} \\ \midrule
Baseline & 4.93 & 77.73 &74.15 \\
token drop & 3.87 (-21.5\%) & 76.58 &72.28 \\ \midrule
\multicolumn{4}{l}{-w/ \textsc{ScTD} (Ours)} \\
\quad $Fi=5$ & 4.69 (-4.9\%) & 78.96 &75.49 \\
\quad $Fi=10$ & 4.25 (-13.8\%) & \textbf{79.58} &\textbf{75.80} \\
\quad $Fi=20$ & 4.04 (-18.1\%) & 78.45 &75.74 \\
\quad $Fi=50$ & 3.92 (-20.5\%) & 79.01 &75.04 \\
\bottomrule
\end{tabular}
}
\caption{Ablation study on different fixed intervals $Fi$ for performing the semantic-align process.}
\label{tab:ablation_fi}
\end{table}
\paragraph{Impact of Fixed Interval $Fi$.} 
In our \textsc{ScTD}, we use a fixed interval $Fi$ to control the frequency for performing the semantic-align process. To verify its impact, we evaluate the performance of \textsc{ScTD} on different $Fi$ and show the results in Table~\ref{tab:ablation_fi}. Observably, too small $Fi$ not only causes much computational overhead, but also affects the stability of hybrid training, thus leading to sub-optimal performance. On the contrary, for the larger $Fi$ (\textit{e.g.}, 50), it may be difficult to make full use of the semantic-consistent learning process, hindering the effect of \textsc{ScTD}. In the case of $Fi=10$, \textsc{ScTD} achieves a better trade-off between costs and performance, which we suggest as the best setting\footnote{Some readers may wonder why the teacher (i.e., model with baseline training) trained with only 1/$Fi$ steps is strong enough to guide the training of student model. One possible reason for this question is that training with hard-to-learn tokens ($Fi$-1) times and training with easy-to-learn tokens once is sufficient to obtain remarkable teacher models, similar to the Lookahead Optimizer~\cite{zhang2019lookahead}, which updates fast weights $k$ times before updating slow weights once.}.

\subsection{Does \textsc{ScTD} indeed alleviate the semantic loss problem?}
\begin{figure}[ht]
\centering
\includegraphics[width=0.47\textwidth]{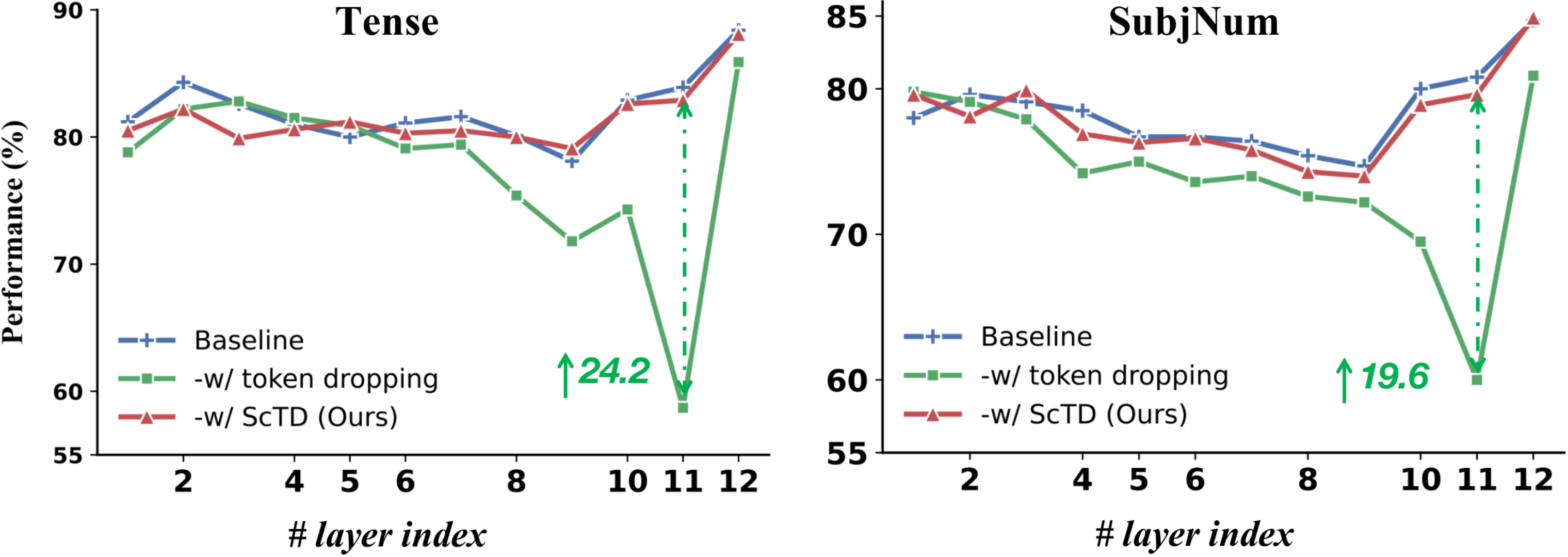}
\caption{The comparison of semantic information on different BERT$\rm_{\texttt{base}}$ layers. We see that \textsc{ScTD} preserves more semantic information than vanilla token dropping.}
\label{fig:analysis_probing}
\end{figure}

Here, we examine whether \textsc{ScTD} can alleviate the limitation of token dropping. Specifically, following the preliminary analyses in~\S\ref{sec:preliminaries}, we compare our \textsc{ScTD} with other counterparts by probing the trained BERT models (as illustrated in Figure~\ref{fig:analysis_probing}) and pertinently evaluating on several semantic-intense tasks (as shown in Table~\ref{tab:analysis_result}). 

\begin{table}[ht]
\centering
\scalebox{0.75}{
\begin{tabular}{lccccc}
\toprule
\multicolumn{1}{c}{\multirow{2}{*}{Method}} & Onto. & CoNLL03 & MRPC & SICK-R &\multicolumn{1}{c}{\multirow{2}{*}{\underline{Avg.}}} \\ \cmidrule(lr){2-5}
\multicolumn{1}{c}{} & \textit{F1} & \textit{F1} & \textit{Acc.} & \textit{Spear.}  \\ \midrule
Token drop & 27.49 & 53.73& 85.50 & 66.16 &\underline{58.22}  \\
\textsc{ScTD} ($\Delta\uparrow$) & \textcolor[RGB]{0,176,80}{\textbf{+2.04}} & \textcolor[RGB]{0,176,80}{\textbf{+2.59}} & \textcolor[RGB]{0,176,80}{\textbf{+1.30}} & \textcolor[RGB]{0,176,80}{\textbf{+2.38}} &\underline{\textcolor[RGB]{0,176,80}{\textbf{+2.08}}} \\
\bottomrule
\end{tabular}
}
\caption{Experimental results of BERT$\rm_{\texttt{base}}$ models on several semantic-intense tasks. We observe that our \textsc{ScTD} brings consistent performance gains.}
\label{tab:analysis_result}
\end{table}

\noindent It can be found that, with our \textsc{ScTD}, BERT learns more semantic information among most layers, especially in dropped layers. Also, \textsc{ScTD} brings consistent and significant performance gains on all semantic-intense tasks against the vanilla token dropping. These results can prove that \textsc{ScTD} is beneficial to address the semantic loss problem.

\section{Related Works}
\label{sec:related}
Pretraining with Transformer-based architectures like BERT~\cite{devlin2019bert} has achieved great success in a variety of NLP tasks~\cite{devlin2019bert,liu2019roberta,he2020deberta,joshi2020spanbert}. Despite its success, BERT-style pretraining usually suffers from unbearable computational expenses~\cite{jiao2020tinybert,zhang2020accelerating}. To this end, several training-efficient approaches are proposed to speed up the pretraining and reduce the computational overhead, such as mixed-precision training~\cite{shoeybi2019megatron}, distributed training~\cite{you2019large}, curriculum learning~\cite{nagatsuka-etal-2021-pre,ding2021progressive} and designing efficient model architectures and optimizers~\cite{gong2019efficient, clark2019electra,zhang2020accelerating,zhang2023mpipemoe,zhong2022improving,sun2023adasam}. These works mainly focus on efficient optimization processes or model architecture changes.

More recently, \citet{hou2022token} propose the token dropping strategy, which exposes a new mode to speed up the BERT pretraining. Without modifying the original BERT architecture or training setting, token dropping is inspired by the dynamic halting algorithm~\cite{dehghani2018universal} and attempts to skip the computations on part of (unimportant) tokens in some middle BERT layers during the forward-propagation process. Owing to its impressive efficiency, token dropping has recently attracted increasing attention~\cite{yao2022random,chiang2022recent}. For instance, \citet{yao2022random} apply the token dropping strategy to broader applications, \textit{e.g.}, both NLP and CV communities.


Along with the line of token dropping, we take a further step by exploring and addressing its limitations. To be specific, we first reveal the semantic loss problem (\S\ref{sec:preliminaries}) in the token dropping, and then propose a novel semantic-consistent learning method (\S\ref{sec:method}) to alleviate this problem and further improve performance and training efficiency. 

\section{Conclusion}
\label{sec:conclusion}
In this paper, we reveal and address the limitation of token dropping in accelerating language model training. Based on a series of preliminary analyses, we find that removing parts of tokens would lead to a semantic loss problem, which causes vulnerable and unstable training. Furthermore, experiments show such a semantic loss will hinder the performance of token dropping in most semantic-intense scenarios. To address this limitation, we improve token dropping with a novel semantic-consistent learning algorithm. It designs two semantic constraints to encourage models to preserve semantic information. Experiments show that our approach consistently and significantly improves downstream performance across all task types and model architectures. In-depth analyses prove that our approach indeed alleviates the problem, and further improves training efficiency.

In future work, we will explore the effectiveness of our method on more advanced discriminative language models~\cite{he2020deberta,zhong2023bag}. Also, it will be interesting to revisit and address the semantic loss problem in efficient training methods for generative language models (such as GPT3~\cite{brown2020language}).

\section*{Limitations}
Our work has several potential limitations.
First, given the limited computational budget, we only validate our \textsc{ScTD} on the Large and Base sizes of BERT models. It will be more convincing if scaling up to the larger model size and applying \textsc{ScTD} to more cutting-edge model architectures. 
On the other hand, besides the downstream performance, we believe that there are still other properties, \textit{e.g.}, generalization and robustness, of MLMs that can be improved by our \textsc{ScTD} approach, which are not fully explored in this work.

\section*{Ethics and Reproducibility Statements}
\paragraph{Ethics} We take ethical considerations very seriously, and strictly adhere to the ACL Ethics Policy. This paper proposes a semantic-consistent algorithm to improve the existing token dropping strategy. The proposed approach aims to speed up the pretraining of BERT-style models, instead of encouraging them to learn privacy knowledge that may cause the ethical problem. Moreover, all pretraining datasets used in this paper are publicly available and have been widely adopted by researchers. Thus, we believe that this research will not pose ethical issues.

\paragraph{Reproducibility} We will publicly release our code in \url{https://github.com/WHU-ZQH/ScTD} and the pretrained models in \url{https://huggingface.co/bert-sctd-base} to help reproduce the experimental results of this paper.

\section*{Acknowledgements}
We are grateful to the anonymous reviewers and the area chair for their insightful comments and suggestions.
This work was supported in part by the National Natural Science Foundation of China under Grants 62225113 and 62076186, and in part by the Science and Technology Major Project of Hubei Province (Next-Generation AI Technologies) under Grant 2019AEA170. Xuebo Liu was supported by Shenzhen Science and Technology Program (Grant No. RCBS20221008093121053). The numerical calculations in this paper have been done on the supercomputing system in the Supercomputing Center of Wuhan University. 

\bibliography{acl2023}
\bibliographystyle{acl_natbib}

\appendix
\section{Appendix}
\label{sec:appendix}
\begin{table*}[h]
\centering
\begin{tabular}{llllcccl}
\toprule
\multicolumn{2}{c}{\textbf{Task}}          & \textbf{\#Train} & \textbf{\#Dev}  & \textbf{\#Class} &\textbf{LR} &\textbf{BSZ} &\textbf{Epochs/Steps} \\ \hline \hline
\multirow{6}{*}{GLUE}       & CoLA      & 8.5K    & 1,042  & 2   &2e-5  &32  &2,668 steps  \\
                            & MRPC        & 3.7K    & 409    & 2   &1e-5  &32  &1,148 steps    \\
                            & STS-B     & 5.7K    & 1,501  & -  &2e-5  &32  &1,799 steps     \\
                            & RTE       & 2.5K    & 278    & 2   &1e-5 &16  &2,036 steps     \\ 
                            & MNLI       & 392K    & 9,815    & 3   &1e-5 &256  &15,484 steps     \\ 
                            & SST-2       & 63.3K    & 873    & 2   &1e-5 &64  &10,467 steps     \\\midrule
\multirow{4}{*}{SuperGLUE}  & BoolQ     & 9.4K    & 3,270  & 2  &1e-5    &16  &10 epochs \\
                            & CB        & 250     & 57     & 2   &2e-5  &16  &20 epochs  \\
                            & MultiRC        & 5.1K      & 953    & 2  &2e-5    &32  &10 epochs \\
                            & COPA        &400     &100    & 2  &2e-5  &16  &10 epochs   \\ \midrule
\multirow{2}{*}{Commonsense QA} & SQuAD v1       &87.6K    &10,570 & -  &3e-5 &12 & 2 epochs      \\
                            & SQuAD v2 &130K    &11,873  & - &3e-5 &12 & 2 epochs    \\
\bottomrule
\end{tabular}
\caption{Data statistics and fine-tuning hyper-parameters of all used tasks in this paper. ``Class'' refers to the label class, ``LR'' means the learning rate and ``BSA'' denotes the batch size. }
\label{appendix_tab_data}
\end{table*}

\subsection{Details of Tasks and Datasets}
\label{appendix_data}
In this work, we conduct extensive experiments on parts of tasks from GLUE and SuperGLUE. In addition, two widely-used commonsense question answering tasks are also used. Here, we introduce the descriptions of the used tasks and datasets in detail. Firstly, we present the statistics of all datasets in Table~\ref{appendix_tab_data}. Then, each task is described as:

\textbf{CoLA} Corpus of Linguistic Acceptability~\cite{warstadt2019neural} is a binary single-sentence classification task to determine whether a given sentence is linguistically ``acceptable''.

\textbf{MRPC} Microsoft Research Paraphrase Corpus~\cite{dolan2005automatically} is a task to predict whether two sentences are semantically equivalent.

\textbf{STS-B} Semantic Textual Similarity~\cite{cer2017semeval} is a task to predict how similar two sentences are on a 1-5 scale in terms of semantic meaning.

\textbf{RTE} Recognizing Textual Entailment~\cite{giampiccolo2007third}, given a premise and a hypothesis, is a task to predict whether the premise entails the hypothesis. 

\textbf{MNLI} The Multi-Genre Natural Language Inference Corpus~\cite{williams2018broad} is a task to predict whether the premise entails the hypothesis, contradicts the hypothesis, or neither, given a premise sentence and a hypothesis sentence.

\textbf{SST-2} The Stanford Sentiment Treebank~\cite{socher2013recursive} is a binary classification task to predict the sentiment of a given
sentence.

\textbf{CB} CommitmentBank~\cite{de2019commitmentbank} is a task that can be framed as three-class textual entailment on a corpus of 1,200 naturally occurring discourses.

\textbf{BoolQ} Boolean Question~\cite{clark2019boolq} is a question answering task where each sample consists of a short passage and a yes/no question about the passage. 

\textbf{MultiRC} Multi-Sentence Reading Comprehension~\cite{khashabi2018looking} is a QA task where each example consists of a context paragraph, a question about that paragraph, and a list of possible answers. The model need to predict which answers are true and which are false. 

\textbf{COPA} Choice of Plausible Alternatives\cite{roemmele2011choice} is a causal reasoning task in which
a system is given a premise sentence and must determine either the cause or effect of the premise
from two possible choices.

\textbf{SQuAD v1} The Stanford Question Answering Dataset~\cite{rajpurkar2016squad} is a popular reading comprehension benchmark, where the answer to each question is a segment of text from the corresponding reading passage.

\textbf{SQuAD v2}  The latest version of
the Stanford Question Answering Dataset~\cite{rajpurkar2018know} is one of the most widely-used reading comprehension benchmarks that require the systems to acquire knowledge reasoning ability.

\subsection{Hyper-parameters of Fine-tuning}
\label{appendix_parameters}
For fine-tuning, we use the BERT models as the backbone PLMs and conduct experiments using the open-source toolkit fairseq\footnote{\url{https://github.com/facebookresearch/fairseq}} and transformers\footnote{\url{https://github.com/huggingface/transformers}}. Notably, we apply the same hyper-parameters to all PLMs for simplicity. The training epochs/steps, batch size, and learning rate for each downstream task are listed in Table~\ref{appendix_tab_data}.

\end{document}